\documentclass[runningheads]{llncs}
\usepackage{multirow}
\usepackage{graphicx}
\usepackage{arydshln}
\usepackage{amsmath}
\usepackage{booktabs}
\usepackage{bm}
\usepackage{amsmath}

\usepackage{amsfonts}
\usepackage{color}

\definecolor{cb_orange}{rgb}{1.0,0.51,0.0}
\definecolor{cb_blue}{rgb}{0.22,0.49,0.72}
\definecolor{cb_green}{rgb}{0.3,0.67,0.29}
\definecolor{cb_red}{rgb}{0.89,0.1,0.11}
\definecolor{cb_pink}{rgb}{1, 0, 0.4}
\definecolor{cb_purple}{rgb}{0.5,0.0,0.5}
\definecolor{cb_cyan}{rgb}{0.0,1.0,1.0}
\definecolor{cb_magenta}{rgb}{1.0,0.0,1.0}

\usepackage{fontawesome} 

\begin{document}
%
%
%
\titlerunning{Clinical-grade Multi-Organ Pathology Report Generation}

\author{
  Jing Wei Tan$^{*1}$, 
  SeungKyu Kim$^{*1}$, 
  Eunsu Kim$^{2}$, 
  Sung Hak Lee$^{2}$, 
  Sangjeong Ahn$^{3}$, 
  Won-Ki Jeong$^{1}$$^($\textsuperscript{\faEnvelopeO}$^)$
}

\authorrunning{Tan et al.}
\def\thefootnote{*}\footnotetext{Equal contribution}
\def\thefootnote{\textsuperscript{\faEnvelopeO}}\footnotetext{Corresponding author: \email{wkjeong@korea.ac.kr}}
%
\institute{$^1$Department of Computer Science and Engineering, College of Informatics, \\ Korea University, Republic of Korea \\
\email{\{jingwei\textunderscore{92},ksk8804,wkjeong\}@korea.ac.kr}\\
$^2$Department of Hospital Pathology, Seoul St. Mary’s Hospital, College of Medicine, The Catholic University of Korea, Republic of Korea \\
\email{kes311az@gmail.com, hakjjang@catholic.ac.kr}\\
$^3$Department of Pathology, Korea University Anam Hospital, College of Medicine, Korea University, Republic of Korea \\
\email{vanitas80@korea.ac.kr}}

%


\title{Clinical-grade Multi-Organ Pathology Report Generation for Multi-scale 
Whole Slide Images via a Semantically Guided Medical Text Foundation Model}
\maketitle              

\begin{abstract}
Vision language models (VLM) have achieved success in both natural language comprehension and image recognition tasks.
However, their use in pathology report generation for whole slide images (WSIs) is still limited due to the huge size of multi-scale WSIs and the high cost of WSI annotation. 
Moreover, in most of the existing research on pathology report generation, sufficient validation regarding clinical efficacy has not been conducted.
%
Herein, we propose a novel 
Patient-level Multi-organ Pathology Report Generation (PMPRG) model
, which utilizes the multi-scale WSI features from our proposed multi-scale regional vision transformer (MR-ViT) model and their real pathology reports to guide VLM training for accurate pathology report generation.
%
%
The model then automatically generates a report based on the provided key features-attended regional features. 
%
We assessed our model using a WSI dataset consisting of 
multiple organs, including the colon and kidney. 
%
Our model achieved a METEOR score of 0.68, demonstrating the effectiveness of our approach. 
This model allows pathologists to efficiently generate pathology reports for patients, regardless of the number of WSIs involved.
The code is available at \url{https://github.com/hvcl/Clinical-grade-Pathology-Report-Generation/tree/main}

\end{abstract}

\section{Introduction}
\label{intro}
The generation of pathology reports is vital, as it offers diagnostic insights derived from the analysis of pathology images (a.k.a WSI), establishing a cornerstone for prognosis evaluation, and ultimately improving diagnostic efficiency for clinicians. 
However, traditional pathology examination requires significant resources, entails labor-intensive procedures, consumes time, and necessitates specialized expertise. 
It may take up to several days to scan and examine slides and generate a pathology report.
Therefore, there exists an increasing demand for automating the report generation process.
%
%

VLMs have gained traction in pathology report generation (PRG).
%
%
Prior works like Zhang \textit{et al.}~\cite{zhang2019pathologist} combined CNN and LSTM for VLM training using ROI or sampled patches from WSIs.
%
PLIP~\cite{huang2023visual}, MI-Zero~\cite{lu2023visual} and CITE~\cite{zhang2023text} adopted CLIP-based training of patch-text pairs, facilitating zero-shot transfer. 
GPC~\cite{nguyen2023gpc} merged CNN with a Transformer-based language model for learning from diverse pathology patches and text pairs. 
%
However, these studies predominantly focus on patch and textual data, requiring specialized expertise for defining ROIs, which is laborious. 
%
Recent approaches in WSI-level report generation include models like Zhang \textit{et al.}~\cite{zhang2020evaluating}, which used CNN and LSTM with thumbnail images and a bag of instances to represent entire WSIs. 
%
Sengupta \textit{et al.}~\cite{sengupta2023automatic} and Guevara \textit{et al.}~\cite{guevara2023caption} utilized pretrained HIPT~\cite{chen2022scaling} for WSI feature extraction and trained with CLIP or LSTM. 
%
%
MI-gen~\cite{MIgen} introduced to insert a Position-Aware Module which adds positional information to each instance into the output obtained through the transformer encoder block.
However, current approaches are limited to generating simple, unstructured descriptions of WSI, often used for biopsy samples, and its clinical efficacy is not yet validated.\\ 
\indent The main motivation of our work stems from the following observations. 
Firstly, unlike the reports generated by most existing work,  
real clinical reports exhibit less descriptive but well-structured formats, varying based on the target organ and disease (see an example report in Fig.~\ref{pipeline}(A).)
%
Existing methods~\cite{zhang2020evaluating,sengupta2023automatic,guevara2023caption} often rely on highly compressed 1D representations from WSIs, raising concerns about their clinical efficacy given the diversity of pathology reports. 
%
As we extend to multi-organ settings, the variability in disease topics intensifies, further emphasizing the importance of addressing this issue.
Secondly, pathologists commonly analyze multiple slides per patient to ensure diagnostic accuracy, necessitating an automatic PRG based on multiple WSIs.
%
Therefore, developing an automatic PRG based on multiple WSIs from a single patient is more closely aligned with clinical practice.
In such a case, the weak-label problem, a commonly found issue with WSI, arises from the predominant assignment of labels at the slide or patient level, and multiple instance learning (MIL) is commonly employed to address this problem (such as DSMIL~\cite{li2021dual} and ZoomMIL~\cite{thandiackal2022differentiable}). 
However, we observed that conventional MIL approaches often prioritize class-specific tasks, such as cancer sub-type classification. 
This emphasis poses a challenge when directly applying MIL to PRG tasks, which encompass a wide array of diagnostic information.
Moreover, we also observed a need to develop an efficient and lightweight encoder that leverages multi-scale information from WSI. 
We empirically found that existing multi-scale WSI encoders are not efficient for training using a large number of WSIs due to the architecture design. \\
\indent In this work, in contrast to traditional methods relying on patches and text, we leverage the multi-scale nature of WSIs along with actual pathology reports to train a VLM tailored for PRG.
%
%
Inspired by HIPT~\cite{chen2022scaling}, which learns features from multiple FoVs on WSIs, we introduce MR-ViT to capture representative features of each multi-scale region. 
%
%
With these regional features, we also propose a Patient-level Multi-organ Pathology Report Generation (PMPRG) model using a medical text foundation model to handle the patients with varying numbers of WSIs, each of different sizes and various scales, making it more suitable for real-world scenarios. 
%
%
The main contributions of our work can be summarized as follows: 
\begin{itemize}
\item 
We propose an efficient unsupervised regional hierarchical model for WSIs, offering improved efficiency in multi-scale feature representation.
By integrating multiple instances from diverse FoVs or multi-scales through a simple two-level feature encoding process, our model captures both local details and broader context effectively. 
This enables the learning of patient-level features alongside text within our model and allows for easy fine-tuning of our report-generation model without compromising quality.
Experimental results validate the superiority of our approach over SOTA multi-scale methods.
%
%
%

\item We propose an automated pathology report generation model that efficiently handles WSIs with various report information 
while allowing easy extension to multiple organs. 
%
%
This approach is suitable for real-world scenarios where
organ attributes vary significantly, and it offers the advantage of generating
clinically valid reports at both patient-level and WSI-level. 
\item To the best of our knowledge, this work is the first attempt to create an explainable 
model that can generate clinical-grade pathology reports, trained using in-house multi-organ datasets consisting of 7422 WSIs with clinical reports in inhomogeneous formats and structures. 
%
%
%

\end{itemize}



\section{Method}
\label{method}
\subsection{Overview and Pre-processing}
\begin{figure}[t!]
\centering
\includegraphics[width=1\columnwidth]{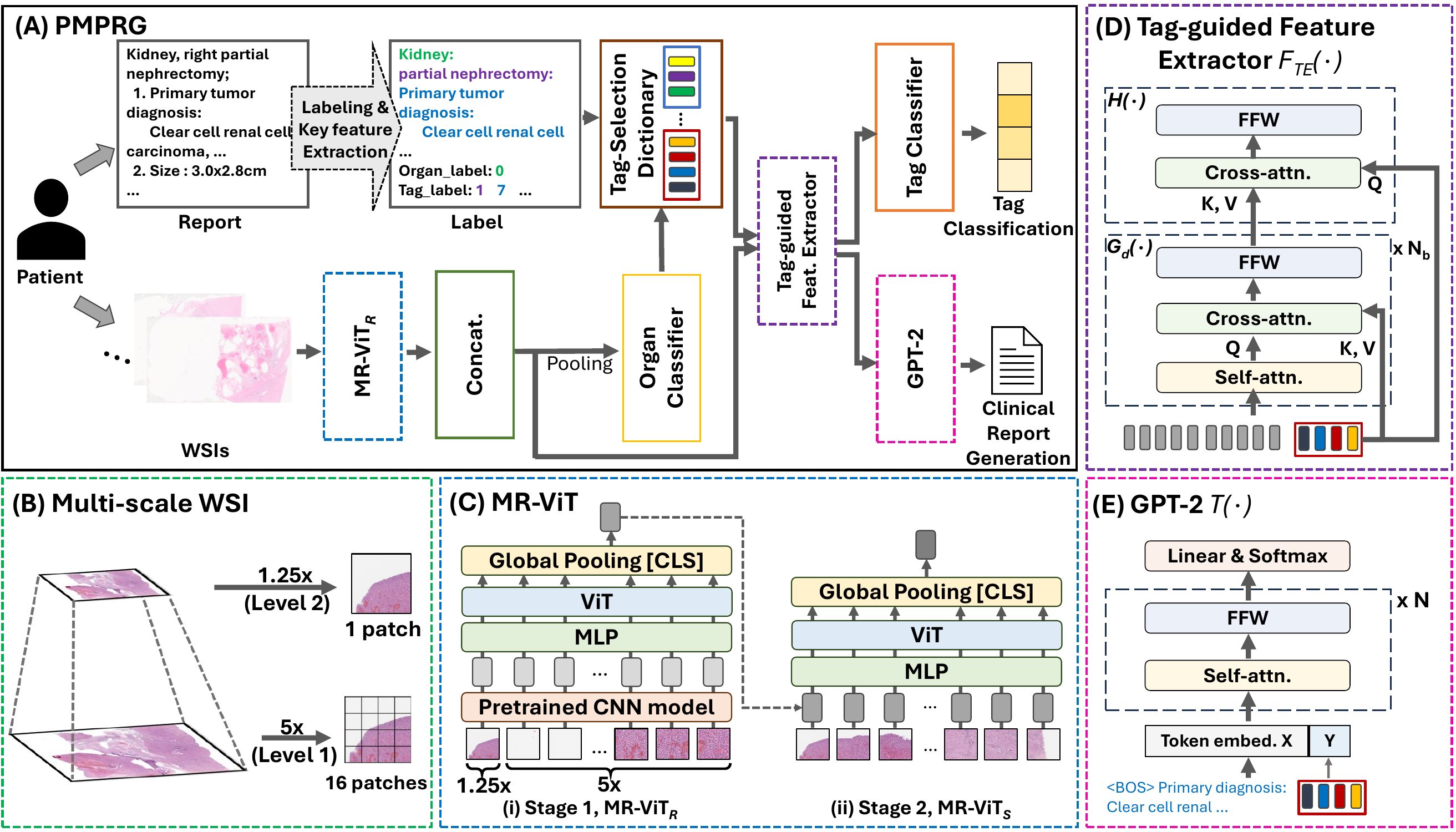}
\vspace{-2.3em}
\caption{Overview of our proposed pipeline: \textbf{(A)}PMPRG, \textbf{(B)}Multi-scale WSI, \textbf{(C)}MR-ViT, \textbf{(D)}Tag-guided feature extractor and \textbf{(E)}GPT-2.}
\label{pipeline}
\vspace{-1.5em}
\end{figure}

%
%
Our proposed model mainly composed of two stages: 
(1) train a multi-scale regional feature extractor in an unsupervised manner and 
(2) train a report generator with the regional feature from the previous stage with reports and labels extracted from reports as shown in Fig.~\ref{pipeline}(C) and (A), respectively.

The input WSIs are first pre-processed by extracting the patches from $1.25\times$ and $5\times$ magnification level in $256\times256\times3$ dimension. 
%
Our dataset includes patients with multiple WSIs, each accompanied by a single pathology report. 
%
%
These reports contain various descriptions for each disease topic (Fig.\ref{pipeline}(A)), varying significantly across organs but showing similarity within the same organ.
%
%
Inspired by radiology report generation methods~\cite{Jing_2018,Tanida_2023,you2022aligntransformer,liu2021exploring,alfarghaly2021automated}, we established predetermined key-tags for each organ in reports, relying exclusively on WSIs guided by pathologists for accurate predictions.
%
%
%
Sentences containing these tags were extracted using a rule-based method.
%
%
If key-tag-related items are missing
, they are excluded from loss calculation. 
We assign tag labels based on diagnosis scenarios after preprocessing, and also each report is labeled with an organ class.
%
The output includes three labels for each report: sentences $Y_{sen} = \{y_{sen_i}\}_{i=1}^{K}$, tag class labels $Y_{tag} = \{y_{tag_i}\}_{i=1}^{K}$, and organ class labels $Y_{org} = \{o \mid o \in \{0, 1, \ldots, n_o\}\}$, where ${K}$ and $n_o$ denote number of tags specific to each organ and total number of organ types within the dataset, respectively. The value of ${K}$ may vary based on the type of organ.
%

%
\subsection{Multi-scale Regional Vision Transformer (MR-ViT)} 
%
%
%
When adapting HIPT~\cite{chen2022scaling} for multi-scale regional representation learning, we introduce two key distinctions from the original approach: (1) utilizing patches of different scales from a specific region as the model input, and (2) reducing computational resources by requiring only two stages to obtain the WSI representation.
In the initial phase of MR-ViT$_{R}$ (Fig.~\ref{pipeline}(C)(i)), 
we construct a 1024×1024 dimensional image representing a specific region $r$ by selecting one patch at level 2 and randomly choosing 15 out of 16 patches at level 1 as shown in Fig.~\ref{pipeline}(B). 
%
%
%
Each region consists of 16 visual tokens ($256 \times 256$), serving as input for ViT. 
Subsequently, we extract features for these visual tokens using a pre-trained model $C$ such as VGG16. 
These extracted embeddings $F_C$ are then utilized in DINO~\cite{dino} to capture multi-scale features for each specific region.
%
%
Following this, the MR-ViT${_R}$ model is employed to extract multi-scale regional representations $F_{R}$ for each WSI. 
$F_{R}$ are then utilized for training MR-ViT${_S}$, as depicted in Fig.~\ref{pipeline}(C)(ii) later on.
We randomly select $Q$ (at $L\times L$ dimension) regions from $F_{R}$ and get $F'_{R}$, which are then fed into the second stage MR-ViT$_{S}$.
At this stage, we aim to capture the multi-scale regional features representing each WSI well in a single vector.  
%
Subsequently, we obtain a WSI representation $F_S$ = MR-ViT$_{S}$($F'_{R}$).
The DINO model hyper-parameter setting is exactly same as HIPT, except for the input size and output size.
%
\subsection{Pathology Report Generation}
As illustrated in Fig.~\ref{pipeline}(A), our model at this stage comprises 6 primary modules:  \textbf{1}) MR-ViT, \textbf{2)} Organ Classifier {\small(${Cls_{org}}$)}, \textbf{3)} Tag-Selection Dictionary, \textbf{4)} Tag-specific Feature Extractor {\small($F_{TE}(\cdot)$)}, \textbf{5)} Tag Classifier {\small(${Cls_{tag}}$)}, 
and \textbf{6)} Conditional Language Model {\small($T(\cdot)$)}.
%

%
\vspace{0.5em}
\textbf{MR-ViT.} All WSIs corresponding to a single patient are patched and processed through the MR-ViT module, generating multiple regional representations.
These representations from multiple WSIs for each patient are then concatenated to form $F_{R,{pat}}$.
%
%
%
%
%

\vspace{0.5em}
\textbf{Organ Prediction.} During batch training, padding is applied along the sequence dimension to $F_{R,{pat}}$,  resulting $F^*_{R,{pat}}$.
Subsequently, after applying average pooling, $F^*_{R,{pat}}$ is being supplied to the organ-classifier ({\small${Cls_{org}}$}), 
consisting of 2 sequential linear layers. 
The organ prediction loss is formulated as follows:
\begin{align*}
L_{org} &= CE Loss(logits_{org}, Y_{org})
\end{align*}

\textbf{Tag-Selection Dictionary.} Given that we have predefined key-tags for each organ, the tags remain consistent for the same organ.
%
Initially, we internally define learnable tag embeddings within the model, denoted as 
$tags_j
=\{{tag_{i} }\}_{i=1}^{K}$, and
$tag\_tot
= {\{tags_j \}_{j=1}^{n_o}}$.
Subsequently, based on the results of organ prediction from the previous module, the corresponding organ's tag-embedding set is retrieved as below:
\begin{align*}
\textit{tag\_batch} \in \mathbb{R}^{B \times K' \times d'} 
&= \{ \text{tags}_{j=\text{Pred}_{org}[k]} \}_{k=1}^{B}
\end{align*}
where $Pred_{org}$ = $Argmax(logits_{org})$.
%
%
%
%

\vspace{0.5em}
\textbf{Tag-specific Feature Extractor.} Tag-specific Feature Extractor ({$F_{TE}(\cdot)$})
comprised of \textit{Decoder Block} ({\small$G_{d}(\cdot)$}) and \textit{Attention Pooler} ({\small$H(\cdot)$}), which takes {\small$tag\_batch$} and $F^{**}_{R,{pat}}$ as inputs for extracting tag-attended visual features ($f_{vt}$).
%
\begin{align*}
f_{vt} &= F_{TE}(tag\_batch, F^{**}_{R,{pat}}) \in \mathbb{R}^{B \times K \times d},\\
&= H(G_{d}(tag\_batch,F^{**}_{R,{pat}}), tag\_batch)
\end{align*}
where 
$F^{**}_{R,{pat}}$ 
represents linear projection of $F^{*}_{R,{pat}}$, aligning dimensions with the tags. 
{\small$F_{TE}(\cdot)$} block follows a standard transformer decoder structure, comprising self-attention, cross-attention, and feed-forward layers. Here, $F^{**}_{R,{pat}}$ is set as the query and {\small$tag\_batch$} as the key and value. 
Similarly, {\small$H(\cdot)$} is composed of cross-attention and feed-forward layers, but does not include a self-attention layer. 
In this case, the query is set as {\small$tag\_batch$} and the key and value are the final output of {\small$G_{d}(\cdot)$}.
After obtaining $f_{vt}$, we reshape it to $f_{vt}'$ $ \in \mathbb{R}^{(B \times K) \times d}$ 
and pass it to {\small${Cls_{tag}}$} which has similar structure to {\small${Cls_{org}}$},
enabling it to make predictions for each tag.
The formulation for tag classifier loss is as follows: 
\begin{align*}
L_{tag} &= CE Loss(logits_{tag}, Y_{tag})
\end{align*}
%
where $logits_{tag}$ represents the tag label prediction.

\vspace{0.5em}
\textbf{Conditional Language Generation.} By feeding the obtained visual features ($f_{vt}'$) into the language model ({\small$T(\cdot)$}),
we enable the model to generate descriptions corresponding to each tag. 
For this task, we employed a pre-trained GPT-2 model~\cite{radford2019language} sourced from PubMed~\cite{papanikolaou2020dare}.
%
By substituting the self-attention layers of GPT-2 with pseudo self-attention (PSA)~\cite{luke2019encoder} layers, we conditioned the language model accordingly. 
Our approach closely follows the methodologies outlined in previous studies [2, 26].
%
%
To preserve the strengths of the foundational language model and for resource-efficient training, we exclusively trained the projection parameters of each block for conditioning input while keeping the remaining parameters frozen. 
Consequently, we derive logits $logits_{sen} = T(f_{vt}', f_l)$ for each sentence in the report.
%
\begin{align*}
L_{sen} &= CE Loss(logits_{sen}, Y_{sen})
\end{align*}
Here, 
$f_l$ represents the token embedding of the report. 
During training, right-shifted $Y_{sen}$ is used as input, and during inference, only the $<BOS>$ token is used.
Finally, the model is trained based on 3 different loss components.
\[
L_{tot} = \alpha L_{org} + \beta L_{tag} + \gamma L_{sen}
\]
where $\alpha$=0.2 , $\beta$=0.6, and $\gamma$=0.2 are weights assigned to each loss component.
\begin{table} [b]
\centering
\setlength{\tabcolsep}{15pt} 
\vspace{-1.3em}
\caption{The performance comparison of patient-level multi-scale image encoder 
with the SOTA methods of multi-scale WSI and an ablation study of different pre-trained models used in MR-ViT.}
\label{image_encoder}
\vspace{-0.8em}
\begin{tabular}{l | c | c} \hline
\multicolumn{1}{c|}{\multirow{2}{*}{Method}} & \multicolumn{2}{c}{Accuracy}   \\ \cline{2-3}
\multicolumn{1}{c|}{}  & \multicolumn{1}{c|}{Diagnosis Type} & \multicolumn{1}{c}{Tumor Grade} \\ \hline
HIPT~\cite{chen2022scaling}  & 0.5455 & 0.3030 \\
ZoomMIL~\cite{thandiackal2022differentiable} & 0.8169 & \textbf{0.4688} \\ \hdashline
MR-ViT (Tan \textit{et al.}~\cite{tan2023multi})   & 0.7273 & 0.3939  \\
MR-ViT (BT~\cite{kang2023benchmarking})  & 0.7879  & 0.3939     \\
MR-ViT (VGG16~\cite{simonyan2014very})   & \textbf{0.8485}   & 0.4242    \\\hline\hline  
\end{tabular}
\vspace{-1em}
\end{table}

\section{Result}
\label{result}


%
%
%
In this study, we collected data from Korea University Anam Hospital covering two organs: the kidney and colon. The dataset includes 1991 patients, with each patient having multiple WSIs, totaling 7422 slides overall. We partitioned the dataset randomly into three subsets: 70\% for training, 20\% for validation, and 10\% for testing.
For PRG evaluation metrics, in addition to the commonly used Natural Language Generation (NLG) metrics, we integrated Clinical Efficacy (CE) metrics to evaluate the model's clinical effectiveness in making predictions. 
For NLG assessment, we utilized three well-established measures: BLEU-n~\cite{BLEU}, METEOR~\cite{METEOR}, and ROUGE-L~\cite{ROUGE}. 
%
We measure F1 score and total accuracy for the CE metric. 
%
%
Implementation details and additional dataset description are provided in the supplementary materials.

\begin{table}[b!]
\centering
\caption{Performance of report generation. For accuracy, we average the individual score of each class (total 26 classes). B-$n$, M, and R-L refers to BLEU, METEOR, and Rouge-L, respectively. Methods with {$^*$} sign indicates generating whole report at once.}
\vspace{-0.8em}
\begin{tabular}{l|cc|cccccc} \hline
\label{maintable}
\multirow{2}{*}{Method} & \multicolumn{2}{c|}{CE metrics} &  \multicolumn{6}{c}{NLG metrics} \\ \cline{2-9}& F1   & Acc.   & B-1    & B-2    & B-3    & B-4    & M      & R-L      \\ \hline
HIPT{$_S^*$} & - & - &    0.1456    &   0.1387     &    0.1305    &    0.1187    &    0.2311    &    0.2880    \\ 
MR-ViT{$_S^*$} & - & -  & 0.3952 & 0.3739 & 0.3518 & 0.3319 & 0.4797 & 0.5137 \\ 
\hline
HIPT{$_S$}     &  0.1808         &    0.2212     &   0.4053   &  0.3589 & 0.3298  &  0.3046  & 0.5381 &  0.4501     \\ 
MR-ViT{$_S$}   &   0.4384   &   0.4285  &  0.4948   & 0.4566 & 0.4274  & 0.3979 &  0.6334 &  0.5449  \\   
\hdashline
HIPT{$_R$}   &   0.1392    &      0.1913     &  0.3866  & 0.3364 &  0.3057  & 0.2819 & 0.5122 & 0.4194         \\
w/o $G_d$, {$Cls_{tag}$}    & -  & -   & 0.5292 & 0.4982 & 0.4739 & 0.4490 & 0.6670 & 0.5888 \\
w/o $G_d$   &   \textbf{0.5786}   &  0.5886 & 0.5362 & 0.5029 & 0.4771 & 0.4500  & 0.6813 &  0.5922 \\
w/o {$Cls_{tag}$}              & -              & -             & 0.5066 & 0.4719 & 0.4444 & 0.4154 & 0.6570 & 0.5747 \\
MR-ViT{$_R$} (Ours)   &  \underline{0.5773}        & \textbf{0.6022}        & \textbf{0.5507} & \textbf{0.5184} & \textbf{0.4925} & \textbf{0.4654} & \textbf{0.6834} & \textbf{0.6033} \\ \hline \hline
\end{tabular}
\vspace{-1em}
\end{table}
\subsection{Experimental Result}
\subsubsection{Image Encoder.}
In this study, we assessed the performance of our proposed MR-ViT by extracting WSI features and classifying them into two specific tasks: (1) diagnosis type and (2) tumor grade.
Given our focus on patient-level rather than slide-level analysis, we obtained slide-level predictions and used the maximum value among the predicted slide classes as the final prediction.
%
We compared our approach with two state-of-the-art (SOTA) multi-resolution WSI methods, HIPT~\cite{chen2022scaling} and ZoomMIL~\cite{thandiackal2022differentiable}. 
Despite attempts to fine-tune the HIPT model for regional feature extraction with our dataset, we encountered significant challenges, including HIPT requiring approximately 300 times longer training time for a single WSI compared to our model under equivalent computational resources. 
%
%
Consequently, we utilized the pre-trained HIPT model directly for feature extraction as in previous studies~\cite{sengupta2023automatic,guevara2023caption}, while training ZoomMIL from scratch in a task-specific manner.
%
%
%
%
%
%
In our experiments (Table~\ref{image_encoder}), MR-ViT with the pre-trained VGG model achieved superior performance compared to both HIPT and ZoomMIL for diagnosis type classification. 
However, ZoomMIL outperformed our MR-ViT and HIPT models for tumor grade classification, likely due to its supervised end-to-end MIL approach whereas MR-ViT and HIPT are trained in unsupervised manner.
%
%
Additionally, we utilized pre-trained encoders such as a pathology image BT model~\cite{kang2023benchmarking}, a multi-scale contrastive encoder~\cite{tan2023multi}, and the ImageNet VGG16 model~\cite{simonyan2014very} to extract patch features for MR-ViT$_{R}$. 
The results indicated that MR-ViT with the pre-trained VGG model showed better performance than the other pre-trained models.
\vspace{-1.2em}
\begin{figure}[t]
\centering
\includegraphics[width=1.0\columnwidth]{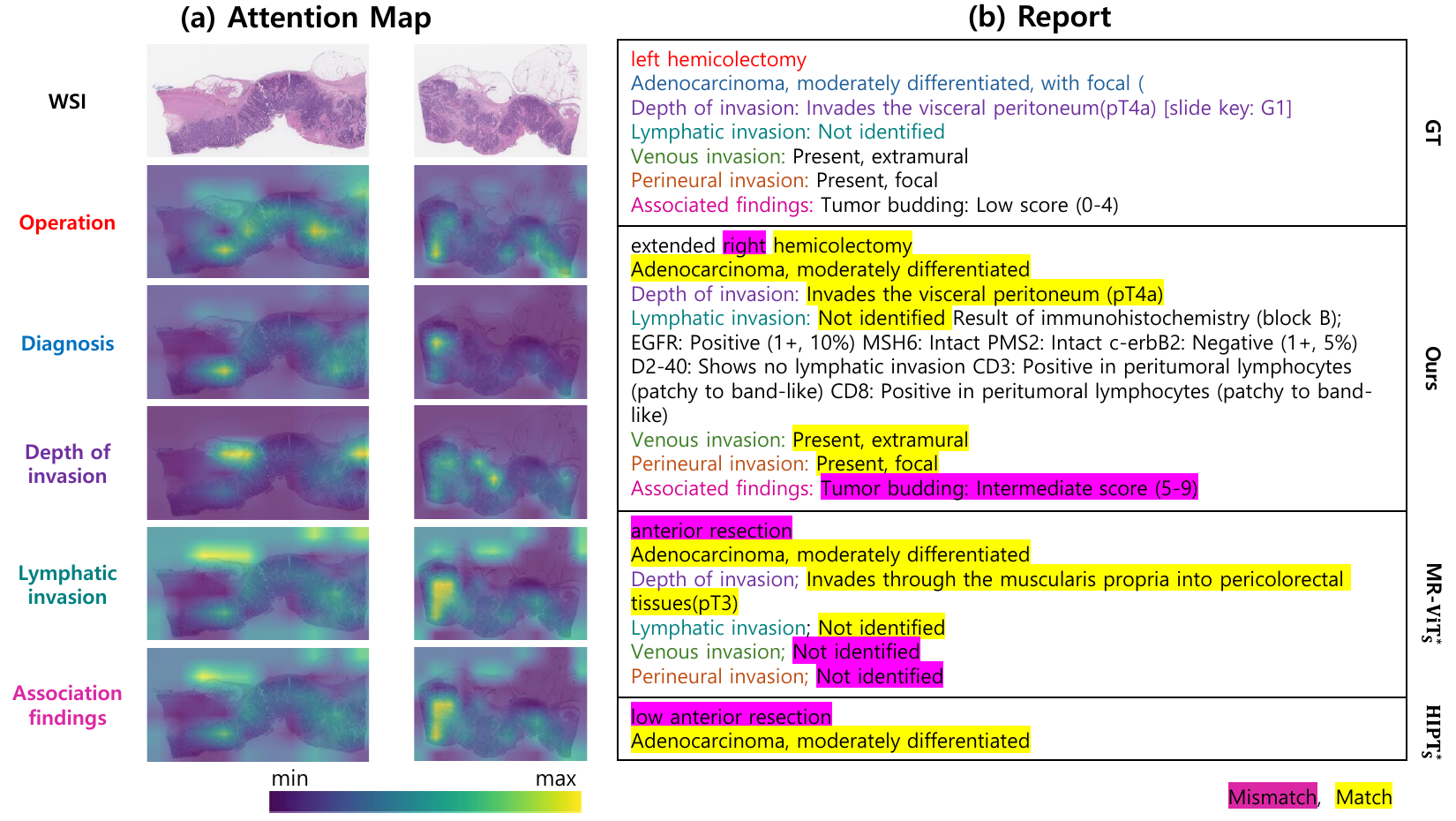}
\vspace{-2.3em}
\caption{
(a) The attention map depicts the importance of WSI regions for specific tags, where brighter regions indicate higher importance. (b) Examples of reports generated using our method and baseline methods.}
\label{temp_visualization}
\vspace{-1mm}
\end{figure} 
\subsubsection{PMPRG.}
We devised three main comparison scenarios for our model:
1) representative visual features extracted using only visual encoder are directly fed into the language model, generating the entire report in one step. For patients with multiple WSIs, we utilized the average-pooled features.
%
%
%
2)follow our overall pipeline but utilizes slide-level features extracted without semantic guidance, and
%
3)follow the approach we proposed.
In each scenario, we compared our encoder (MR-ViT) with HIPT. Due to computational resource constraints, HIPT~\cite{chen2022scaling} was utilized with pre-trained weights without further training, following default settings. 
%
Table~\ref{maintable} presents comprehensive results for report generation, highlighting the effectiveness of segmenting the report into distinct tags and transforming visual features into specific vectors for each tag, compared to directly relaying features from the visual encoder to the language model.  
%
%
This approach, especially for HIPT, results in a notable improvement in METEOR scores(approximately 0.3 increase from HIPT{$_S^*$} to HIPT{$_S$}), enhancing report generation quality. 
%
Moreover, providing semantic guidance for each tag during the feature compression stage significantly contributes to generating high-quality reports compared to using slide-level features extracted without guidance (MR-ViT{$_S$} $\rightarrow$ MR-ViT{$_R$}), notably improving CE metric performance (approximately 17\% more accurate predictions).
%
%
Although HIPT exhibits a decrease in both CE and NLG metrics, it is speculated that ineffective feature extraction from the beginning may have contributed to limited impact on the results, as evidenced by notably low scores achieved by HIPT.
Additionally, an ablation study for each module reveals that each module contributes to guiding the model's report generation favorably.
Specifically, the  {\small$Cls_{tag}$} module aids in generating clinically accurate descriptions for each tag, resulting in improvements of 0.03 in both METEOR and Rouge-L scores, along with an increase of approximately 0.05 in BLEU scores. \\
\indent Fig. 2(a) illustrates qualitative results with attention maps for interpreting the model's predictions. WSIs displayed are from a single patient, with the model integrating information from multiple WSIs for a comprehensive diagnosis. Attention weights are obtained from {\small$H(\cdot)$} and yield individual attention maps for each tag, revealing where the model focuses its attention and providing insights into specific regions considered during diagnosis. 
%
Fig. 2(b) showcases the generated reports from each model. Generating the entire report at once (HIPT{$_S^*$}, MR-ViT{$_S^*$}) may lack descriptions for tags present in the actual report. Conversely, our model includes all tags, which seems reasonable considering the consistent nature of pathology reports within the same organ. 
More results are provided in supplementary material.

\section{Conclusion}
\label{conclusion}
%
%
%
In this study, we present a streamlined multi-scale regional encoder for WSIs and an innovative patient-level multi-organ pathology generation model. Our proposed image encoder enhances training efficiency for multi-scale WSI images compared to the baseline method. Moreover, by harnessing the multi-scale regional features, our pathology report generation model can provide pathologists with practical, real-world pathology reports. Future plans include extending our model to incorporate three magnification scales and incorporating more data from diverse organs.


\subsubsection{Acknowledgements.}
This study was approved by the institutional review board of Korea University Anam Hospital (IRB NO.2024AN0190). This work was partially supported by the National Research Foundation of Korea (RS-2024-00349697, NRF-2021R1A6A1A13044830), the Institute for Information \& Communications Technology Planning \& Evaluation (IITP-2024-2020-0-01819), the Technology development Program(RS-2024-00437796) funded by the Ministry of SMEs and Startups(MSS, Korea), the Korea Health Industry Development Institute (RS-2021-KH113146) and a Korea University Grant.

\subsubsection{Disclosure of Interests.} 
There are no conflicts of interest to declare.

\bibliographystyle{splncs04}
\bibliography{Paper-1738}

\end{document}


%
\clearpage
\section*{Supplementary}
\label{supplementary}
Loss Function Details

\setcounter{table}{0}
\renewcommand{\thetable}{S\arabic{table}}
\setcounter{figure}{0}
\renewcommand{\thefigure}{S\arabic{figure}}
\vspace{-3em}
\begin{figure}[htb]
\includegraphics[width=0.99\linewidth]{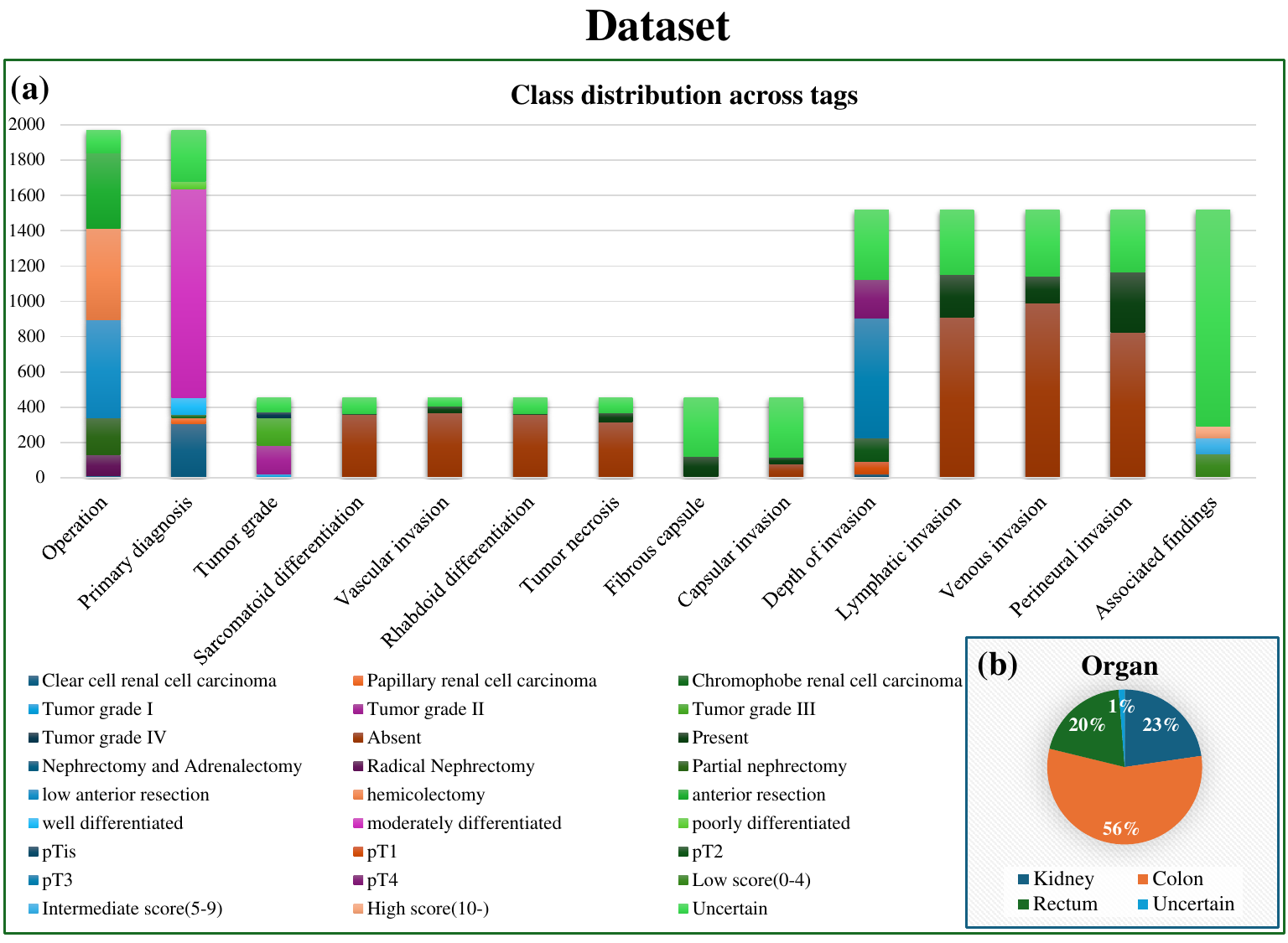}
\vspace{-1em}
\caption{Dataset composition. 
(a) Distribution of inner classes for each tag. The x-axis represents the tag names, while the y-axis indicates the number of occurrences for each tag. After selecting inner classes for each tag, we integrated them into one group. Therefore, the output dimension of the tag classifier becomes the size of that group (27: 26 classes $+$ 1 'uncertain' class).
(b) Distribution of organs in the dataset. Data labeled as 'uncertain' is extracted from colon or rectum, where there are no keywords related to organs in the reports.
}
\label{sup_dataset}
\vspace{-3em}
\end{figure}
\begin{table}[htb!]
\setlength{\tabcolsep}{20pt}
\centering
\caption{Implementation details. For training stage-1, we adopt the default settings from HIPT. To evaluate its effectiveness, we employ a classifier consisting of two linear layers followed by a softmax layer. While for stage-2 training, we run for 300 epochs and choose the model with the best tag classification accuracy on the validation set as the representative model. For comparative experiments without a {\small ${Cls_{tag}}$}, we select the model with the smallest $L_{sen}$.}
\begin{tabular}{l|c|c} \hline
\multicolumn{1}{c|}{} & \multicolumn{1}{c|}{stage-1} & \multicolumn{1}{c}{stage-2} \\ \hline
Epoch Number         & \multicolumn{1}{c|}{500}     & \multicolumn{1}{c}{300}     \\
Batch Size           & 64                          & 1                           \\
Optimizer            & SGD                         & AdamW                       \\
Learning Rate        & 1e-1                        & 3e-3                       \\\hline\hline
\end{tabular}
\vspace{-3em}
\end{table}


\begin{figure}[h]
\includegraphics[width=0.99\linewidth]{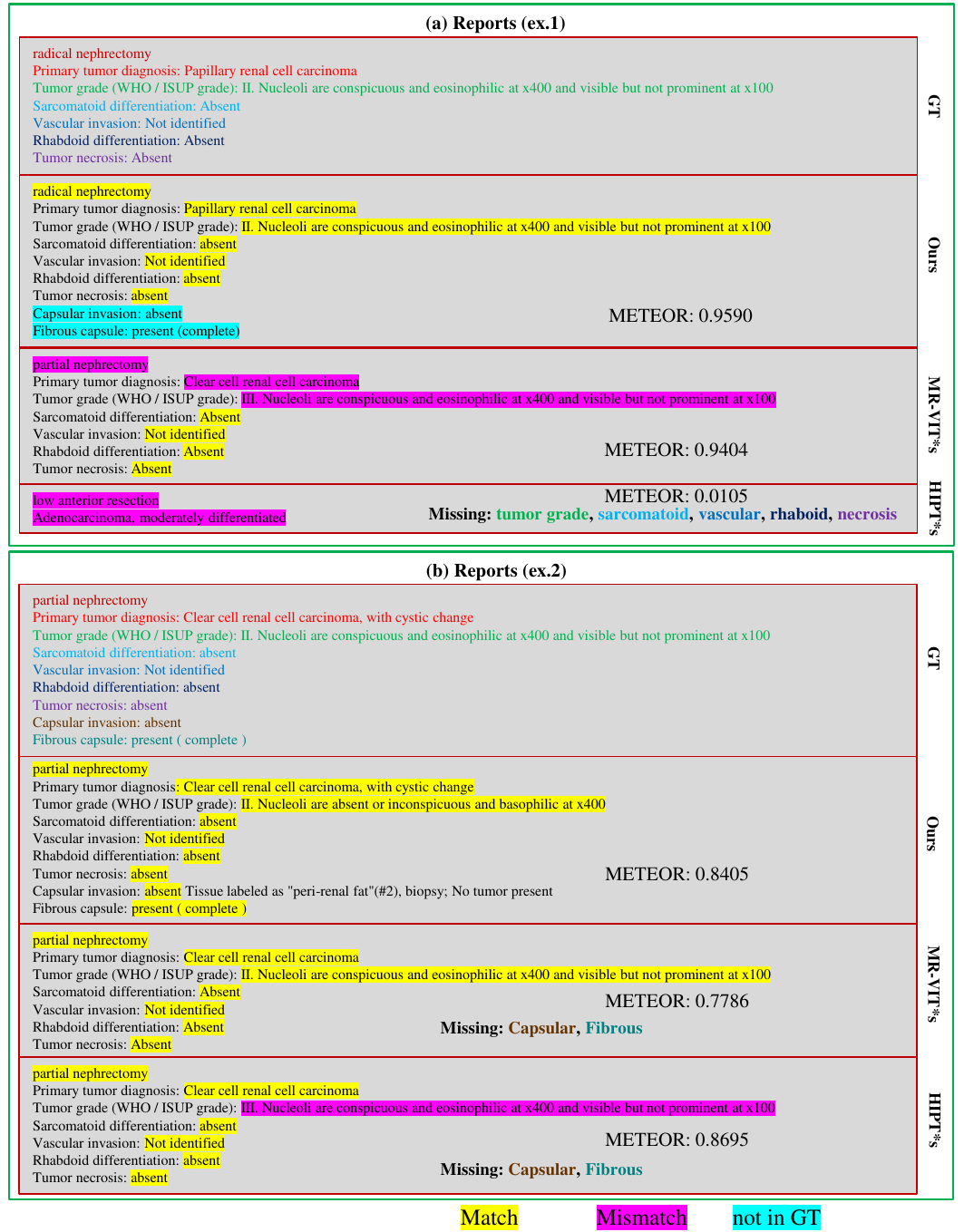}
\caption{
Additional report generation results for the kidney. 
(a) Ours provides accurate descriptions for all tags, generating additional descriptions for tags not mentioned in the ground truth report. Despite the high number of incorrect descriptions in MR-ViT{$_S^*$}, our METEOR score is only slightly upper than MR-ViT{$_S^*$}.
(b) While our approach provides accurate descriptions for all tags, HIPT{$_S^*$} has one incorrect description and misses descriptions for two tags. Surprisingly, the METEOR score is higher for HIPT{$_S^*$} than Ours. This result suggests that NLG metrics may not absolutely represent the quality of medical report generation.
}
\label{sup_dataset}
\end{figure}